\documentclass[final]{cvpr}

\usepackage{times}
\usepackage{epsfig}
\usepackage{graphicx}
\usepackage{amsmath}
\usepackage{amssymb}

\usepackage{booktabs}
\usepackage{multirow}
\usepackage{siunitx}
\usepackage{xcolor,colortbl}
\usepackage[numbers, sort]{natbib}

\definecolor{mygray}{gray}{0.93}

\usepackage[pagebackref=true,breaklinks=true,colorlinks,bookmarks=false]{hyperref}

\def\cvprPaperID{88886666} 
\def\confYear{CVPR 2021}

\begin{document}


\title{Tracking Instances as Queries}

\author{Shusheng Yang$^{1,2}$\thanks{Equal contributions. This work was done while Shusheng Yang was interning at Applied Research Center (ARC), Tencent PCG.}, \ \ Yuxin Fang$^{1*}$, \ \ Xinggang Wang$^{1}$\thanks{Corresponding author, E-mail: {\tt xgwang@hust.edu.cn.}}, \ \ Yu Li$^{2}$, \\
\vspace{0.15cm}
Ying Shan$^{2}$, \ \ Bin Feng$^{1}$, \ \ Wenyu Liu$^{1}$ \\
$^1$School of EIC, Huazhong University of Science \& Technology \\
$^2$Applied Research Center (ARC), Tencent PCG\\
}

\maketitle

\begin{abstract}
Recently, query based deep networks catch lots of attention owing to their end-to-end pipeline and competitive results on several fundamental computer vision tasks, such as object detection~\cite{SparseRCNN}, semantic segmentation~\cite{SETR}, and instance segmentation~\cite{QueryInst}. However, how to establish a query based video instance segmentation (VIS) framework with elegant architecture and strong performance remains to be settled. In this paper, we present \textbf{QueryTrack} (i.e., tracking instances as queries), a unified query based VIS framework fully leveraging the intrinsic one-to-one correspondence between instances and queries in QueryInst~\cite{QueryInst}. The proposed method obtains 52.7 / 52.3 AP on YouTube-VIS-2019 / 2021 datasets, which wins the 2-nd place in the YouTube-VIS Challenge at CVPR 2021 \textbf{with a single online end-to-end model, single scale testing \& modest amount of training data}. We also provide QueryTrack-ResNet-50 baseline results on YouTube-VIS-2021 val set as references for the VIS community.
\end{abstract}

\section{Introduction}

 Video Instance Segmentation (VIS) ~\cite{MaskTrackR-CNN} is an emerging computer vision task and get rapid development since it was proposed.
 This task extends the traditional instance segmentation to the temporal domain and requires detecting, classifying, segmenting, and tracking visual instances simultaneously in the given videos.
 Similar to other video based tasks like Video Object Segmentation~\cite{GCMVOS, MGST-VOS} and Video Object Detection~\cite{DSFNet}, video instance segmentation provides a natural understanding of video scenes.
 Achieving accurate and robust video instance segmentation in real-world scenarios can greatly promote the development of video analysis.
 
 Given the inherent relationship between video instance segmentation and instance segmentation, prevalent video instance segmentation methods~\cite{MaskTrackR-CNN, SipMask, CrossVIS, MaskProp, CompFeat, STEm-Seg} prefer utilizing off-the-shelf instance segmentation approaches with various modules for inter-frame feature aggregation and temporal instances association.
 As a result, modern video instance segmentation methods usually use the one-to-many matching between predictions and ground truth instances, thus the inference process is sensitive to manual-designed post-process operators and far from end-to-end.
 Moreover, to associate instances across video frames, current VIS methods~\cite{MaskTrackR-CNN, SipMask} require heuristic association approach and bring lots of hyper-parameters.

 To remedy these issues, we propose a unified end-to-end query based video instance segmentation method, termed as QueryTrack (\textit{i.e., tracking instances as queries}).
 The proposed method is built upon the state-of-the-art query based instance segmentation method QueryInst~\cite{QueryInst}, which exploits the intrinsic one-to-one correspondence in object queries across different stages, and one-to-one correspondence between mask RoI features and object queries in the same stage.
 Moreover, an elaborate tracking head is introduced to fully leverage the potential of instance queries for the temporal association.
 With the one-to-one correspondence between instances and queries, QueryTrack enjoys a unified end-to-end paradigm.
 Moreover, the well-designed tracking head greatly reduces the number of hyper-parameters.
 
 The proposed QueryTrack is evaluated on the YouTube-VIS Challenge $2021$, which achieves $52.3$\% AP on the $\mathtt{test}$ set and the $2^{nd}$ place on the final leaderboard. 
 We also conduct experiments on YouTube-VIS $2019$~\cite{MaskTrackR-CNN} dataset, where QueryTrack outperforms a deal of previous state-of-the-art methods.
 To facilitate future research, we also report results of QueryInst powered by ResNet-$50$~\cite{ResNet} backbone on YouTube-VIS-$2021$ $\mathtt{val}$ set\footnote{It is common sense that the widely used YouTube-VIS $2019$ benchmark usually suffers from \textbf{high variances}~\cite{CrossVIS}.
 The YouTube-VIS $2021$ benchmark is an augmented version of the $2019$ version that producing much more reliable results (we observe $\sim 0.5$ AP noise for a wide range of VIS models).
 We encourage the community to evaluate VIS methods on the $2021$ version for better reproducibilities.}.
 With a simple framework and competitive performances, we hope QueryTrack can serve as a strong baseline for video instance segmentation.

\section{Method}

\begin{figure*}
 \centering
 \includegraphics[width=2.0\columnwidth]{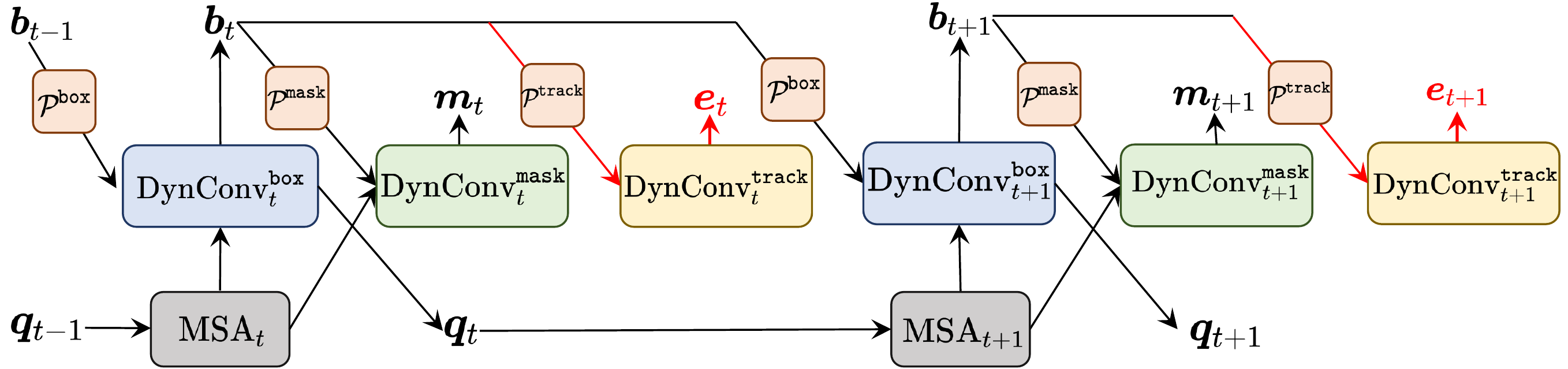}
 \caption{Overall pipeline of QueryTrack. The black arrows indicate the original pipeline of QueryInst~\cite{QueryInst}, while the red arrows stand for the introduced track pipelines to tackle the video instance segmentation problem.}
 \label{fig: QueryTrack}
 \end{figure*}

 In this section, we explicate the architecture design of QueryTrack in detail. Fig~\ref{fig: QueryTrack} gives an overall illustration of the proposed methods.

\subsection{Query Based Instance Segmentation}

 As aforementioned, QueryTrack is built on the top of QueryInst~\cite{QueryInst}, the well-designed query based instance segmentation framework.
 The overall object detection and instance segmentation pipelines are summarized as follows.
 
 \noindent
 \textbf{Object Detection.}
 The object detection pipeline can be formulated as:
 \begin{equation}
 \begin{aligned}
 \boldsymbol{x}_{t}^{\mathtt{box}} \leftarrow & \ \mathcal{P}^{\mathtt{box}}\left(\boldsymbol{x}^{\mathtt{FPN}}, \boldsymbol{b}_{t - 1}\right), \\
 \boldsymbol{q}_{t - 1}^* \leftarrow & \ \mathrm{MSA}_{t}\left(\boldsymbol{q}_{t - 1} \right), \\
 \boldsymbol{x}_{t}^{\mathtt{box*}}, \boldsymbol{q}_{t} \leftarrow & \ \mathrm{DynConv}^{\mathtt{box}}_{t} \left(\boldsymbol{x}_{t}^{\mathtt{box}}, \boldsymbol{q}_{t - 1}^* \right), \\
 \boldsymbol{b}_{t} \leftarrow & \ \mathcal{B}_{t}\left(\boldsymbol{x}_{t}^{\mathtt{box*}}\right),
 \end{aligned}
 \end{equation}
 \noindent
 $\boldsymbol{q} \in \mathbf{R}^{N \times d} $ indicates the instance query while $N$ and $d$ denote the total number and dimension of instance query, respectively.
 For bounding box prediction, at stage $t$, a pooling operator $\mathcal{P}^{\mathtt{box}}$ extracts the current stage bounding box feature $\boldsymbol{x}_{t}^{\mathtt{box}}$ from FPN feature $\boldsymbol{x}^{\mathtt{FPN}}$ under the guidance of previous stage bounding box prediction $\boldsymbol{b}_{t - 1}$.
 Meanwhile, a multi-head self-attention module $\mathrm{MSA}_{t}$ is applied to the input query $\boldsymbol{q}_{t - 1}$ to get the transformed query $\boldsymbol{q}_{t - 1}^*$.
 Then, a box dynamic convolution module $\mathrm{DynConv}^{\mathtt{box}}_{t}$ takes $\boldsymbol{x}_{t}^{\mathtt{box}}$ and $\boldsymbol{q}_{t - 1}^*$ as inputs and enhances the $\boldsymbol{x}_{t}^{\mathtt{box}}$ by reading $\boldsymbol{q}_{t - 1}^*$ and generates $\boldsymbol{q}_{t}$ for the next stage. 
 Finally, the enhanced bounding box feature $\boldsymbol{x}_{t}^{\mathtt{box}*}$ are fed into the box prediction branch $\mathcal{B}_{t}$ for current stage bounding box prediction $\boldsymbol{b}_{t}$.

 \noindent
 \textbf{Instance Segmentation.}
 For instance mask prediction, a region-wise pooling operator $\mathcal{P}^{\mathtt{mask}}$ extracts the current stage mask feature $\boldsymbol{x}_{t}^{\mathtt{mask}}$ from FPN feature $\boldsymbol{x}^{\mathtt{FPN}}$, under the guidance of current stage bounding box prediction $\boldsymbol{x}_{t}$.
 A mask dynamic convolution module $\mathrm{DynConv}^{\mathtt{mask}}_{t}$ enhances the original mask feature $\boldsymbol{x}_{t}^{\mathtt{mask}}$ and generates $\boldsymbol{x}_{t}^{\mathtt{mask*}}$.
 Afterwards, current stage mask head $\mathcal{M}_{t}$ generates the instance level mask prediction $\boldsymbol{m}_{t}$ by a stack of convolutional layers. The overall procedure of instance mask generation can be formulated as follows:
 \begin{equation}
 \begin{aligned}
 \boldsymbol{x}_{t}^{\mathtt{mask}} \leftarrow & \ \mathcal{P}^{\mathtt{mask}}\left(\boldsymbol{x}^{\mathtt{FPN}}, \boldsymbol{b}_{t}\right), \\
 \boldsymbol{x}_{t}^{\mathtt{mask*}} \leftarrow & \ \mathrm{DynConv}^{\mathtt{mask}}_{t} \left(\boldsymbol{x}_{t}^{\mathtt{mask}}, \boldsymbol{q}_{t - 1}^* \right), \\
 \boldsymbol{m}_{t} \leftarrow & \ \mathcal{M}_{t}\left(\boldsymbol{x}_{t}^{\mathtt{mask*}}\right),
 \end{aligned}
 \end{equation}

 \noindent
 \textbf{Bipartite Matching.}
 Following ~\cite{QueryInst, YOLOS}, we adapt hungarian matching to build the one-to-one correspondences between predictions and ground truth instances. The matching cost of Hungarian matcher is defined as:
 \begin{equation}
 \mathcal{L}_{Hungarian} = \lambda_{cls} \cdot \mathcal{L}_{cls}+\lambda_{L1} \cdot \mathcal{L}_{L1}+\lambda_{giou} \cdot \mathcal{L}_{giou}
 \end{equation}
 where $\mathcal{L}_{cls}$, $\mathcal{L}_{L1}$ and $\mathcal{L}_{giou}$ indicate the focal loss, L1 loss and generalized $\mathtt{IoU}$ loss, respectively. $\lambda_{cls}$, $\lambda_{L1}$ and $\lambda_{giou}$ are set as the same as ~\cite{QueryInst}.

\subsection{Contrastive Tracking Head}

 \noindent
 \textbf{Dynamic Instance Embedding.}
 To perform temporal instances association, we first embed all instances to a latent space by a dynamic instance embedding head.
 Specifically, the embedding process can be formulated as follows:

 \begin{equation}
 \begin{aligned}
 \boldsymbol{x}_{t}^{\mathtt{track}} \leftarrow & \ \mathcal{P}^{\mathtt{track}}\left(\boldsymbol{x}^{\mathtt{FPN}}, \boldsymbol{b}_{t}\right), \\
 \boldsymbol{x}_{t}^{\mathtt{track*}} \leftarrow & \ \mathrm{DynConv}^{\mathtt{track}}_{t} \left(\boldsymbol{x}_{t}^{\mathtt{track}}, \boldsymbol{q}_{t - 1}^* \right), \\
 \boldsymbol{e}_{t} \leftarrow & \ \mathcal{T}_{t}\left(\boldsymbol{x}_{t}^{\mathtt{track*}}\right),
 \end{aligned}
 \end{equation}
 Similar to mask prediction, firstly, a region-wise pooling operator extracts instance feature $\boldsymbol{x}_{t}^{\mathtt{track}}$, a track dynamic convolution module $\mathrm{DynConv}^{\mathtt{track}}_{t}$ enhances the instance feature under the guidance of instance query. Then, a linear projection module $\mathcal{T}_{t}$ projects $\boldsymbol{x}_{t}^{\mathtt{track*}}$ to a latent space and generates instance emebdding $\boldsymbol{e}_{t}$.

 \noindent
 \textbf{Contrastive Learning.}
 Following ~\cite{MaskTrackR-CNN, CrossVIS}, we take a pair of frames as inputs to train the tracking head.
 During training, the frame pairs are randomly sampled from a training video.
 One of the frames is picked as the key frame, which is fed to the instance segmentation network to get a set of instance predictions.
 While the other frame is treated as a reference frame, which aims to provide ground truth identities and reference instance embeddings.
 Assuming there is a detected instances $\mathcal{I}_{i}$ at the key frame, and there are $N$ already identified instances in the reference frame.
 It's clear that there is at most one existing identity in reference frame can be assigned to the detected instances.
 The probability of assigning label $n$ to detected instance $\mathcal{I}_{i}$ can be formulated as:
 \begin{equation}
 p_i(n) = 
 \begin{cases}
 \dfrac{\exp{(\mathbf{e}_{i}^{\top}\mathbf{e}_n})}{1 + \sum_{j = 1}^{N}
 \exp{(\mathbf{e}_{i}^{\top}\mathbf{e}_j})} & \mathrm{if \ } n \in [1, N],\\ 
 & \\
 \dfrac{1}{1 + \sum_{j = 1}^{N}
 \exp{(\mathbf{e}_{i}^{\top}\mathbf{e}_j})} & \mathrm{otherwise},
 \end{cases}
 \end{equation}
 where $\mathbf{e}_{i}$ and $\mathbf{e}_{j}$ denote the instance embedding of $\mathcal{I}_{i}$ and $n$ instance embeddings in reference frame.
 Different from ~\cite{MaskTrackR-CNN}, which introduces a cross-entropy loss function to optimize the tracking head, QueryTrack adapts a contrastive focal loss to reduce the conflict of multi-task learning.
 Specifically, the loss function for tracking heads is defined as follow:
 \begin{equation}
 \begin{aligned}
 p_{i}^{*}(n)=
 \begin{cases}
 p_i(n) & \mathrm{if \ } \mathcal{I}_{i}=\mathcal{I}_{n}, \\
 & \\
 1-p_i(n) & \mathrm{otherwise},
 \end{cases}
 \end{aligned}
 \end{equation}
 \begin{equation}
 \mathcal{L}_{track} = - \alpha_t (1 - p_i^*(n))^{\gamma}\log(p_i^*(n)),
 \end{equation}

\subsection{Online Instance Association}

 Tracking instances across video frames purely based on the instance embedding is non-trivial as appearance similarity might be confused by instance deformation, occlussion and background change. Similar to ~\cite{MaskTrackR-CNN, FairMOT}, QueryTrack leverages several tracking clues such as spatial similarity, detection confidence and category consistency to perform better instance association. Specifically, assume there are $M$ candidate instances and $N$ already identified instances, the matching factor between one candidate instance $m \in [1, M]$ one identified instance $n \in [1, N]$ can be formulated as:
 \begin{equation}
 \mathcal{F}_{m,n} = \mathcal{S}_{m,n} \cdot \frac{1+\mathtt{IoU}(\mathtt{b}_{m}, \mathtt{b}_{n})}{2} \cdot \frac{1+\pi_{m}}{2} \cdot \delta(\mathtt{c}_m, \mathtt{c}_n)
 \end{equation}
 where $\mathtt{IoU}(\mathtt{b}_{m}, \mathtt{b}_{n})$ indicates the bounding box $\mathtt{IoU}$ (intersection over union) between candidate instance $m$ and identified instance $n$, $\pi_{m}$ indicates the detection confidence of candidate instance $m$, and $\delta(\mathtt{c}_m, \mathtt{c}_n)$ is an indicator function which gets $1$ when the two instances have the same category predictions ($\mathtt{c}_m=\mathtt{c}_n$) and gets $0$ otherwise. $\mathcal{S}_{m,n}$ indicates the normalized appearance similarity between two instances. Specifically, the similarity is normalized by a bi-directional softmax~\cite{Quasi-Dense}, the computation process can be formulated as follows:
 \begin{equation}
 \mathcal{S}_{m,n}=\left(\frac{\exp{(\mathbf{e}_{m}^{\top} \mathbf{e}_{n}})}{\sum_{k=1}^{N}{\exp{(\mathbf{e}_{m}^{\top} \mathbf{e}_{k}})}}+\frac{\exp{(\mathbf{e}_{m}^{\top} \mathbf{e}_{n}})}{\sum_{k=1}^{M}{\exp{(\mathbf{e}_{k}^{\top} \mathbf{e}_{n}})}}\right) / 2
 \end{equation}

\section{Experiments}

\subsection{Datasets}

 We mainly evaluate QueryTrack on YouTube-VIS $2021$ dataset, which is uesd for YouTube-VIS Challenge $2021$.
 Besides, we also report the system level comparisons between QueryTrack and several state-of-the-art methods on YouTube-VIS $2019$ dataset.



\subsection{Implementation Details}
 \label{details}
 \noindent
 \textbf{Training Setup.}
 The basic training setup of QueryTrack is mainly following the original QueryInst~\cite{QueryInst}.
 Specifically, the R-CNN head of QueryTrack contains $6$ stages and the total number of queries is set to $300$.
 We adopt the recently proposed transformer network~\cite{Swin, msgtransformer} as the backbone, and use COCO pre-trained weights for parameter initialization.
 The training process on YouTube-VIS consists of $12$ epochs in total.
 For each iter, the batch size is set to $32$ and we use AdamW optimizer with an initial learning rate of $1.25\times10^{-5}$.
 The learning rate decreases by $10$ at $9^{th}$ and $11^{th}$ epoch.
 Data augmentation includes random flip, multi-scale input, and random crop. Input images are resized such that the shorter side is at least $320$ and at most $800$, while the other side no longer than $1333$.
 QueryTrack only exploit modest amount of training data.
 Unlike other methods in VIS competitions, we \textbf{do not use} Open-Iamge data~\cite{OpenImage} for better performance.

 \noindent
 \textbf{Inference.}
 Since most of the videos in both YouTube-VIS $2019$~\cite{MaskTrackR-CNN} and YouTube-VIS $2021$ have no more than $10$ video instances, during inference we only extract the top $10$ instance predictions as valid candidates.
 The instance masks are generated from the final stage mask head, and the final stage tracking head is used to associate temporal instances.
 All input images during the inference stage are resized to have their shorter side being $640$ and their longer side no longer than $1333$.

 \begin{table}[t]
 \centering
 \setlength{\tabcolsep}{3.4 pt}
 \begin{tabular}{l|c|ccc}
 \hline

 \hline

 \hline

\rowcolor{mygray}
 Method
 & Backbone
 & AP
 & AP$_{50}$
 & AP$_{75}$
 \\

 \hline
 \hline

 MaskTrack R-CNN~\cite{MaskTrackR-CNN}
 & \multirow{5}{*}{\shortstack[c]{ResNet-$50$}}
 & $28.6$
 & $48.9$
 & $29.6$
 \\

 SipMask-VIS~\cite{SipMask}
 & 
 & $31.7$
 & $52.5$
 & $34.0$
 \\

 CrossVIS~\cite{CrossVIS}
 & 
 & $34.2$
 & $54.4$
 & $37.9$
 \\

 IFC~\cite{IFC-VIS}
 &
 & $36.6$
 & $57.9$
 & $39.3$
 \\

 \textbf{QueryTrack (Ours)}
 &
 & $\mathbf{38.8}$
 & $\mathbf{62.0}$
 & $\mathbf{42.3}$
 \\

 \hline

 \hline

 \hline
 
 \end{tabular}
 \smallskip
 \caption{QueryInst with ResNet-$50$ backbone baseline results on YouTube-VIS-$2021$ $\mathtt{val}$ set.
 Our results are highlight in \textbf{bold}.}
 \label{tab: baseline}
\end{table}

\subsection{YouTube-VIS-2021 Val Set Baseline}
We report QueryInst-ResNet-$50$ baseline results on the YouTube-VIS-$2021$ $\mathtt{val}$ set as references for the VIS community.
The training setting keeps the same as described in Sec.~\ref{details}, except the number of queries is $100$ (which will degenerate the performance), and we don't use crop augmentation.
During inference, the input video frame resolution is $360 \times 640$ (\ie, $360\mathrm{p}$).

The results are shown in Tab.~\ref{tab: baseline}.
We demonstrate that QueryTrack with ResNet-$50$ backbone can serve as a very strong baseline for video instance segmentation future research.

\subsection{Main Results}

 Tab.~\ref{tab: leaderboard} shows the results in the final leaderboard of YouTube-VIS Challenge $2021$.
 With a single model, our QueryTrack achieves $52.3$ AP in the $\mathtt{test}$ set of YouTube-VIS $2021$, and wins the $2^{nd}$ place in YouTube-VIS Challenge $2021$.

 Tab.~\ref{tab: systemlevel comparison} shows the system level comparisons between QueryTrack and state-of-the-art video instance segmentation methods.
 As shown in the table, QueryTrack outperforms previous state-of-the-art video instance segmentation methods by a large margin.

\section{Conclusion}

We present QueryTrack, a unified query based end-to-end framework for the challenging video instance segmentation task.
We build our method upon the state-of-the-art instance segmentation model QueryInst~\cite{QueryInst} with an elaborate tracking head.
Despite the simple \& concise framework, QueryTrack produces very strong results and achieves the $2^{nd}$ place in the YouTube-VIS Challenge $2021$.
We also provide some competitive baselines on YouTube-VIS-$2021$ $\mathtt{val}$ set to facilitate future research, and we encourage the community to evaluate VIS models on YouTube-VIS-$2021$ dataset for better reproducibilities.

\begin{table}[t]
 \centering
 \setlength{\tabcolsep}{3.6 pt}
 \begin{tabular}{l|ccccc}
 \hline

 \hline

 \hline

\rowcolor{mygray}
 Team
 & AP
 & AP$_{50}$
 & AP$_{75}$
 & AR$_{1}$
 & AR$_{10}$
 \\

 \hline
 \hline

 tuantng~\cite{nguyen20211st}
 & $54.1$
 & $74.2$
 & $61.1$
 & $43.3$
 & $58.9$
 \\

 \textbf{QueryTrack (Ours)}
 & $\mathbf{52.3}$
 & $\mathbf{76.7}$
 & $\mathbf{57.7}$
 & $\mathbf{43.9}$
 & $\mathbf{57.0}$
 \\

 vidit$98$~\cite{MSN}
 & $49.1$
 & $68.1$
 & $54.5$
 & $41.0$
 & $55.0$
 \\

 linhj~\cite{SeqMaskR-CNN}
 & $47.8$
 & $69.3$
 & $52.7$
 & $42.2$
 & $59.1$
 \\

 hongsong.wang
 & $47.6$
 & $68.4$
 & $52.9$
 & $41.4$
 & $54.6$
 \\

 gb7~\cite{MaskProp}
 & $47.3$
 & $66.5$
 & $51.5$
 & $40.5$
 & $51.6$
 \\

 zfonemore
 & $46.1$
 & $64.4$
 & $51.0$
 & $38.3$
 & $50.6$
 \\

 DeepBlueAI
 & $46.0$
 & $64.6$
 & $52.0$
 & $38.7$
 & $54.2$
 \\

 zhangxuan
 & $41.0$
 & $62.0$
 & $42.9$
 & $37.3$
 & $47.1$
 \\

 Suqi.lmh
 & $32.3$
 & $48.8$
 & $36.2$
 & $30.2$
 & $38.2$
 \\

 \hline

 \hline

 \hline
 \end{tabular}
 \smallskip
 \caption{Results in the YouTube-VIS Challenge $2021$, compared to top $10$ other participants. Our results are highlighted in \textbf{bold}.}
 \label{tab: leaderboard}
\end{table}

\begin{table}[t]
 \centering
 \setlength{\tabcolsep}{5. pt}
 \begin{tabular}{l|ccc}
 \hline

 \hline

 \hline

\rowcolor{mygray}
 Method
 & AP
 & AP$_{50}$
 & AP$_{75}$
 \\

 \hline
 \hline

 MaskTrack R-CNN~\cite{MaskTrackR-CNN}
 & $30.3$
 & $51.1$
 & $32.6$
 \\

 SipMask-VIS~\cite{SipMask}
 & $33.7$
 & $54.1$
 & $35.8$
 \\

 STEm-Seg~\cite{STEm-Seg}
 & $34.6$
 & $55.8$
 & $37.9$
 \\

 CompFeat~\cite{CompFeat}
 & $35.3$
 & $56.0$
 & $38.6$
 \\

 CrossVIS~\cite{CrossVIS}
 & $36.6$
 & $57.3$
 & $39.7$
 \\

 VisTR~\cite{VisTR}
 & $40.1$
 & $64.0$
 & $45.0$
 \\

 IFC~\cite{IFC-VIS}
 & $44.6$
 & $69.2$
 & $49.5$
 \\

 MaskProp~\cite{MaskProp}
 & $46.6$
 & $51.2$
 & $-$
 \\

 SeqMask R-CNN~\cite{SeqMaskR-CNN}
 & $47.6$
 & $71.6$
 & $51.8$
 \\
 
 \hline

  \textbf{QueryTrack} (X-$101$-DCN~\cite{ResNeXt})
  & $\mathbf{48.8}$
  & $\mathbf{74.2}$
  & $\mathbf{52.4}$
  \\

 \textbf{QueryTrack} (Swin-L~\cite{Swin})
 & $\mathbf{52.7}$
 & $\mathbf{78.9}$
 & $\mathbf{57.9}$
 \\

 \hline

 \hline

 \hline
 \end{tabular}
 \smallskip
 \caption{Comparisons with state of the art methods on YouTube-VIS $2019$ dataset. Our results are highlighted in \textbf{bold}.}
 \label{tab: systemlevel comparison}
\end{table}

{\small
\bibliographystyle{ieee_fullname}
\bibliography{egbib}

\begin{thebibliography}{10}\itemsep=-1pt

\bibitem{STEm-Seg}
Ali Athar, S. Mahadevan, Aljosa Osep, L. Leal-Taix{\'e}, and B. Leibe.
\newblock Stem-seg: Spatio-temporal embeddings for instance segmentation in
  videos.
\newblock In {\em ECCV}, 2020.

\bibitem{MaskProp}
Gedas Bertasius and Lorenzo Torresani.
\newblock Classifying, segmenting, and tracking object instances in video with
  mask propagation.
\newblock In {\em CVPR}, 2020.

\bibitem{SipMask}
Jiale Cao, Rao~Muhammad Anwer, Hisham Cholakkal, Fahad~Shahbaz Khan, Yanwei
  Pang, and Ling Shao.
\newblock Sipmask: Spatial information preservation for fast image and video
  instance segmentation.
\newblock In {\em ECCV}, 2020.

\bibitem{msgtransformer}
Jiemin Fang, Lingxi Xie, Xinggang Wang, Xiaopeng Zhang, Wenyu Liu, and Qi Tian.
\newblock Msg-transformer: Exchanging local spatial information by manipulating
  messenger tokens.
\newblock {\em arXiv:2105.15168}, 2021.

\bibitem{YOLOS}
Yuxin Fang, Bencheng Liao, Xinggang Wang, Jiemin Fang, Jiyang Qi, Rui Wu,
  Jianwei Niu, and Wenyu Liu.
\newblock You only look at one sequence: Rethinking transformer in vision
  through object detection.
\newblock {\em arXiv preprint arXiv:2106.00666}, 2021.

\bibitem{QueryInst}
Yuxin Fang, Shusheng Yang, Xinggang Wang, Yu Li, Chen Fang, Ying Shan, Bin
  Feng, and Wenyu Liu.
\newblock Instances as queries.
\newblock {\em arXiv preprint arXiv:2105.01928}, 2021.

\bibitem{CompFeat}
Yang Fu, Linjie Yang, Ding Liu, Thomas~S Huang, and Humphrey Shi.
\newblock Compfeat: Comprehensive feature aggregation for video instance
  segmentation.
\newblock {\em arXiv preprint arXiv:2012.03400}, 2020.

\bibitem{MSN}
Vidit Goel, Jiachen Li, Shubhika Garg, Harsh Maheshwari, and Humphrey Shi.
\newblock Msn: Efficient online mask selection network for video instance
  segmentation.
\newblock {\em arXiv:2106.10452}, 2021.

\bibitem{ResNet}
Kaiming He, Xiangyu Zhang, Shaoqing Ren, and Jian Sun.
\newblock Deep residual learning for image recognition.
\newblock In {\em CVPR}, 2016.

\bibitem{IFC-VIS}
Sukjun Hwang, Miran Heo, Seoung~Wug Oh, and Seon~Joo Kim.
\newblock Video instance segmentation using inter-frame communication
  transformers.
\newblock {\em arXiv preprint arXiv:2106.03299}, 2021.

\bibitem{OpenImage}
Alina Kuznetsova, Hassan Rom, Neil Alldrin, Jasper Uijlings, Ivan Krasin, Jordi
  Pont-Tuset, Shahab Kamali, Stefan Popov, Matteo Malloci, Alexander
  Kolesnikov, et~al.
\newblock The open images dataset v4.
\newblock {\em IJCV}, 2020.

\bibitem{GCMVOS}
Yu Li, Zhuoran Shen, and Ying Shan.
\newblock Fast video object segmentation using the global context module.
\newblock In {\em ECCV}, 2020.

\bibitem{SeqMaskR-CNN}
Huaijia Lin, Ruizheng Wu, Shu Liu, Jiangbo Lu, and Jiaya Jia.
\newblock Video instance segmentation with a propose-reduce paradigm.
\newblock {\em arXiv preprint arXiv:2103.13746}, 2021.

\bibitem{DSFNet}
Lijian Lin, Haosheng Chen, Honglun Zhang, Jun Liang, Yu Li, Ying Shan, and
  Hanzi Wang.
\newblock Dual semantic fusion network for video object detection.
\newblock In {\em ACMMM}, 2020.

\bibitem{Swin}
Ze Liu, Yutong Lin, Yue Cao, Han Hu, Yixuan Wei, Zheng Zhang, Stephen Lin, and
  Baining Guo.
\newblock Swin transformer: Hierarchical vision transformer using shifted
  windows.
\newblock {\em arXiv preprint arXiv:2103.14030}, 2021.

\bibitem{nguyen20211st}
Thuy~C Nguyen, Tuan~N Tang, Nam~LH Phan, Chuong~H Nguyen, Masayuki Yamazaki,
  and Masao Yamanaka.
\newblock 1st place solution for youtubevos challenge 2021: Video instance
  segmentation.
\newblock {\em arXiv preprint arXiv:2106.06649}, 2021.

\bibitem{Quasi-Dense}
Jiangmiao Pang, Linlu Qiu, Xia Li, Haofeng Chen, Qi Li, Trevor Darrell, and
  Fisher Yu.
\newblock Quasi-dense similarity learning for multiple object tracking.
\newblock {\em arXiv preprint arXiv:2006.06664}, 2020.

\bibitem{SparseRCNN}
Peize Sun, Rufeng Zhang, Yi Jiang, Tao Kong, Chenfeng Xu, Wei Zhan, Masayoshi
  Tomizuka, Lei Li, Zehuan Yuan, Changhu Wang, et~al.
\newblock Sparse r-cnn: End-to-end object detection with learnable proposals.
\newblock In {\em CVPR}, 2021.

\bibitem{VisTR}
Yuqing Wang, Zhaoliang Xu, Xinlong Wang, Chunhua Shen, Baoshan Cheng, Hao Shen,
  and Huaxia Xia.
\newblock End-to-end video instance segmentation with transformers.
\newblock {\em arXiv preprint arXiv:2011.14503}, 2020.

\bibitem{ResNeXt}
Saining Xie, Ross Girshick, Piotr Doll{\'a}r, Zhuowen Tu, and Kaiming He.
\newblock Aggregated residual transformations for deep neural networks.
\newblock In {\em CVPR}, 2017.

\bibitem{MaskTrackR-CNN}
Linjie Yang, Yuchen Fan, and Ning Xu.
\newblock Video instance segmentation.
\newblock In {\em ICCV}, 2019.

\bibitem{CrossVIS}
Shusheng Yang, Yuxin Fang, Xinggang Wang, Yu Li, Chen Fang, Ying Shan, Bin
  Feng, and Wenyu Liu.
\newblock Crossover learning for fast online video instance segmentation.
\newblock {\em arXiv preprint arXiv:2104.05970}, 2021.

\bibitem{FairMOT}
Yifu Zhang, Chunyu Wang, Xinggang Wang, Wenjun Zeng, and Wenyu Liu.
\newblock Fairmot: On the fairness of detection and re-identification in
  multiple object tracking.
\newblock {\em arXiv preprint arXiv:2004.01888}, 2020.

\bibitem{SETR}
Sixiao Zheng, Jiachen Lu, Hengshuang Zhao, Xiatian Zhu, Zekun Luo, Yabiao Wang,
  Yanwei Fu, Jianfeng Feng, Tao Xiang, Philip~HS Torr, et~al.
\newblock Rethinking semantic segmentation from a sequence-to-sequence
  perspective with transformers.
\newblock In {\em CVPR}, 2021.

\bibitem{MGST-VOS}
Qiang Zhou, Zilong Huang, Lichao Huang, Yongchao Gong, Han Shen, Wenyu Liu, and
  Xinggang Wang.
\newblock Motion-guided spatial time attention for video object segmentation.
\newblock In {\em ICCVW}, 2019.

\end{thebibliography}
}

\end{document}